\pdfoutput=1

\documentclass[11pt]{article}

\usepackage{acl}
\usepackage{amssymb}
\usepackage{times}
\usepackage{latexsym}

\usepackage[T1]{fontenc}

\usepackage[utf8]{inputenc}

\usepackage{microtype}

\usepackage{inconsolata}
\usepackage{tabularx}
\usepackage{graphicx}
\usepackage{amsmath}
\usepackage{algorithm}
\usepackage{algpseudocode}
\usepackage{booktabs}
\usepackage{multirow}
\usepackage{CJKutf8}
\usepackage{longtable}
\usepackage{subcaption}
\usepackage{float}
\usepackage{enumitem}
\title{TIM: A Large-Scale Dataset and large Timeline Intelligence Model for Open-domain Timeline Summarization}

\author{
  \begin{tabular}{ccccc}
    \textbf{Chuanrui Hu\thanks{These authors contributed equally.}\textsuperscript{1}} &
    \textbf{Wei Hu\thanks{These authors contributed equally.}\textsuperscript{1}} &
    \textbf{Penghang Yu\textsuperscript{2}} &
    \textbf{Hua Zhang\textsuperscript{1}}  &
    \textbf{Bing-Kun Bao\thanks{Corresponding author.}\textsuperscript{2}}
  \end{tabular} \\[1ex]
  \begin{tabular}{c}
    \textsuperscript{1}Qihoo360 \\
    \textsuperscript{2}Nanjing University of Posts and Telecommunications, Nanjing, China \\
  \end{tabular} \\[1ex]
  \begin{tabular}{cccc}
    huchuanrui@360.cn & huwei5@360.cn & 2022010201@njupt.edu.cn & zhanghua3@360.cn \\
    \multicolumn{4}{c}{bingkunbao@njupt.edu.cn}
  \end{tabular}
}

\begin{document}
\begin{CJK}{UTF8}{gbsn}
\maketitle

\begin{abstract}
Open-domain Timeline Summarization (TLS) is crucial for monitoring the evolution of news topics. To identify changes in news topics, existing methods typically employ general Large Language Models (LLMs) to summarize relevant timestamps from retrieved news. While general LLMs demonstrate capabilities in zero-shot news summarization and timestamp localization, they struggle with assessing topic relevance and understanding topic evolution. Consequently, the summarized information often includes irrelevant details or inaccurate timestamps.
To address these issues, we propose the first large \textbf{T}imeline \textbf{I}ntelligence \textbf{M}odel (\textbf{TIM}) for open-domain TLS, which is capable of effectively summarizing open-domain timelines. Specifically, we begin by presenting a large-scale TLS dataset, comprising over 1,000 news topics and more than 3,000 annotated TLS instances. Furthermore, we propose a progressive optimization strategy, which gradually enhance summarization performance. It employs instruction tuning to enhance summarization and topic-irrelevant information filtering capabilities. Following this, it exploits a novel dual-alignment reward learning method that incorporates both semantic and temporal perspectives, thereby improving the understanding of topic evolution principles. Through this progressive optimization strategy, TIM demonstrates a robust ability to summarize open-domain timelines. Extensive experiments in open-domain demonstrate the effectiveness of our TIM.
\end{abstract}

\section{Introduction}
With the exponential growth of news driven by the global expansion of the Internet, information has become increasingly fragmented and overwhelming. As a consequence, users face challenges in discerning key moments within a given topic. In the context of long-term developments or shifts in public opinion, the public often seeks to grasp key moments from complex information \cite{zhao2016generating}. To address this need, the timeline summarization (\textbf{TLS}) task was proposed \cite{yu2021multi,hu2024moments,qorib2025just}. It is designed to extract and organize such information chronologically, aiding users in identifying  events and their progression over time \cite{tran2015timeline}.

\begin{figure}[t]
  \includegraphics[width=\columnwidth]{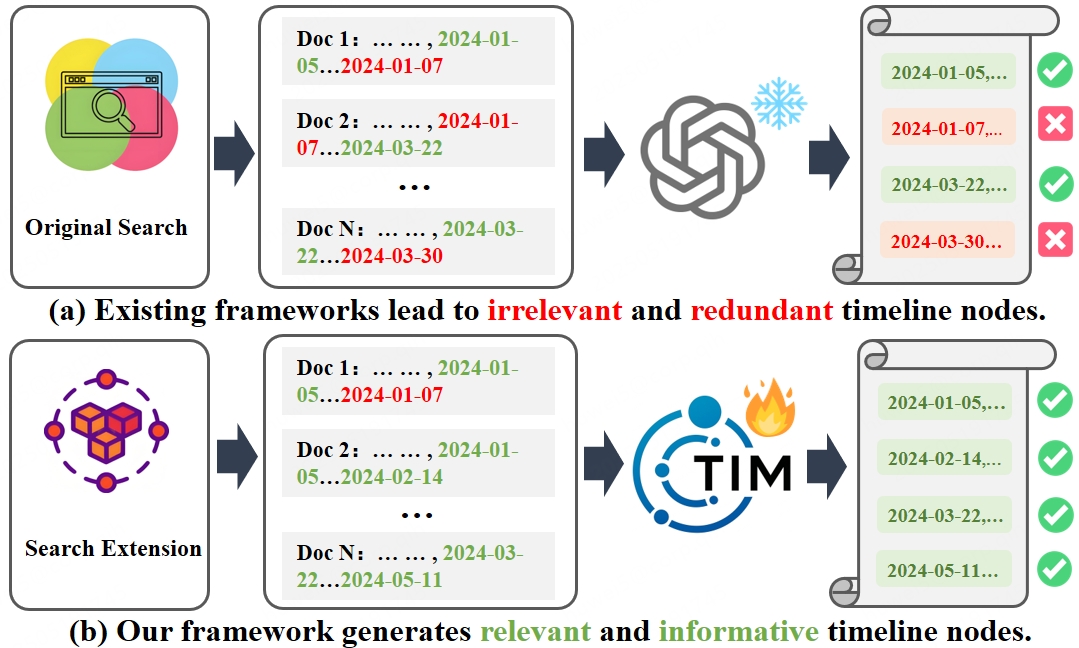}
  \caption{Illustration comparing existing frameworks with our framework. (a) Existing methods fail to filter irrelevant content, leading to redundant TLS. (b) Our method effectively suppresses irrelevant content, yielding more concise and relevant TLS. }
  \label{fig:1}
  \vspace{-15pt}
\end{figure}

Recognizing that previous TLS tasks \cite{hu2024moments}—often designed for specific domains—lacked generalization capabilities, which limited their applicability in real-world environments. \cite{wu2025unfolding} propose the Open-domain TLS task. This task integrates Retrieval Augmented Generation (RAG) with TLS by retrieving news through a search engine, and subsequently utilizing a general Large Language Model (LLM) for summarization. Leveraging the LLM's abilities for zero-shot news summarization and timestamp localization, their efforts have yielded initial success in the field of open domain TLS.


\begin{table*}[t]
\centering
\resizebox{\textwidth}{!}{%
\begin{tabular}{l c c c c c| c c c c}
\hline
\textbf{Dataset} & language & \# of topics & \# of timelines & \# of articles & \# of domains & Avg. \# of articles & Avg. duration (days) & Avg. \( l \) & Avg. \( k \) \\
\hline
\textbf{T17} & EN & 9 & 19  & 9,653 & 1 & 508 & 212 & 36 & 2.9 \\
\textbf{Crisis} & EN & 4 & 22  & 50,820 & 1  & 2310 & 343 & 29 & 1.3 \\
\textbf{Entities} & EN & 47 & 47  & 45,073 & 1  & 959 & 4437 & 23 & 1.2 \\
\textbf{OPEN-TLS} & EN & 50 & 50  & - & 5  & - & 4139 & 23 & 1.8  \\
\textbf{CNTLS} & ZH & 77 & 77 & 43,428  &  5 &  564 & 55 & 6 & 1.4 \\
\hline
\textbf{TLS-\uppercase\expandafter{\romannumeral1}} & ZH & \textbf{1189} & \textbf{3567}  & \textbf{1,252,017} & \textbf{12}  & 1053 & 1309 & 17 & 1.6 \\
\hline
\end{tabular}%
}

\caption{Statistics of the TLS-I datasets. The left side of the figure presents the core characteristics of our dataset, highlighting its large scale and broad domain coverage. The right side provides standard statistical information to support overall understanding. Avg. \( l \) represents the average number of sentences per topic, and Avg. \( k \) represents the average summary length of a single timeline.}\label{tab:1}
\vspace{-10pt}
\end{table*}

Despite their initial success, general LLMs struggle with assessing topic relevance and understanding topic evolution. This issue becomes especially pronounced in real-world narratives, where temporal cues and content description are often vague or ambiguous \cite{song2024temporal}. Consequently, the summarized results often include irrelevant details or inaccurate timestamps (see Section Table~\ref{tab:2} for experimental verification). Therefore, it is necessary to develop a large timeline intelligent model for TLS tasks, that can accurately identify key timestamps while filtering out extraneous information. Such a model must be capable of discerning the relationships between content and topics, as well as tracking the evolution of these topics. However, the development of such models is severely hindered by a lack of data, as most available datasets contain fewer than 100 news timelines sourced from a limited number of domains \cite{binh2013predicting,tran2015timeline}. Consequently, there is a pressing need for the construction of a large-scale dataset and an effective method to develop large timeline intelligent models for TLS.

To address these issues, we construct the first large-scale TLS dataset: TLS-\uppercase\expandafter{\romannumeral1}, comprising over 1,000 topics, 3,000 annotated timelines, and 100,000 news articles. Compared to existing datasets, our dataset covers broader domains, richer temporal information, and high-quality human annotations, making training a large timeline intelligence model for TLS feasible. Furthermore, we propose a progressive optimization strategy, that first enhances summarization and topic-irrelevant information filtering capabilities, followed by an exploration of the principles of topic evolution. Specifically, we employ a topic-aware sampling strategy to construct a challenging training set for instruction tuning. By utilizing samples with high relevance to the topic—containing content closely associated with the topic—the model can rapidly develop summarization skills. Conversely, samples with low relevance, which include substantial unrelated content, enhance the model's ability to perceive topics. 
Following this, we design a dual-alignment reward learning method to deepen the model's understanding of topic evolution patterns. This method encourages the model to consider both the semantic alignment of the generated summaries with reference summaries, and the alignment of temporal timestamp, enabling the model to perceive topic evolution more effectively.
Through this progressive optimization strategy, we develop the first large \textbf{T}imeline \textbf{I}ntelligence \textbf{M}odel (\textbf{TIM}) capable of effectively summarizing open-domain timelines. Our dataset, codes and trained models will be made publicly available upon paper acceptance.

Our main contributions are as follows:
\begin{itemize}[itemsep=0pt, topsep=0pt]
\item We introduce and open-source the first large-scale TLS dataset. This dataset makes training a large timeline intelligence model feasible.

\item We propose a novel progressive optimization strategy, which enhances the understanding of topic relevance and topic evolution.

\item Experimental results show that our model outperforms general LLMs on open-domain TLS. Additionally, performance improves consistently with model size, further validating the effectiveness of our dataset.

\end{itemize}

\begin{figure*}[t]
  \centering
  \includegraphics[width=\linewidth]{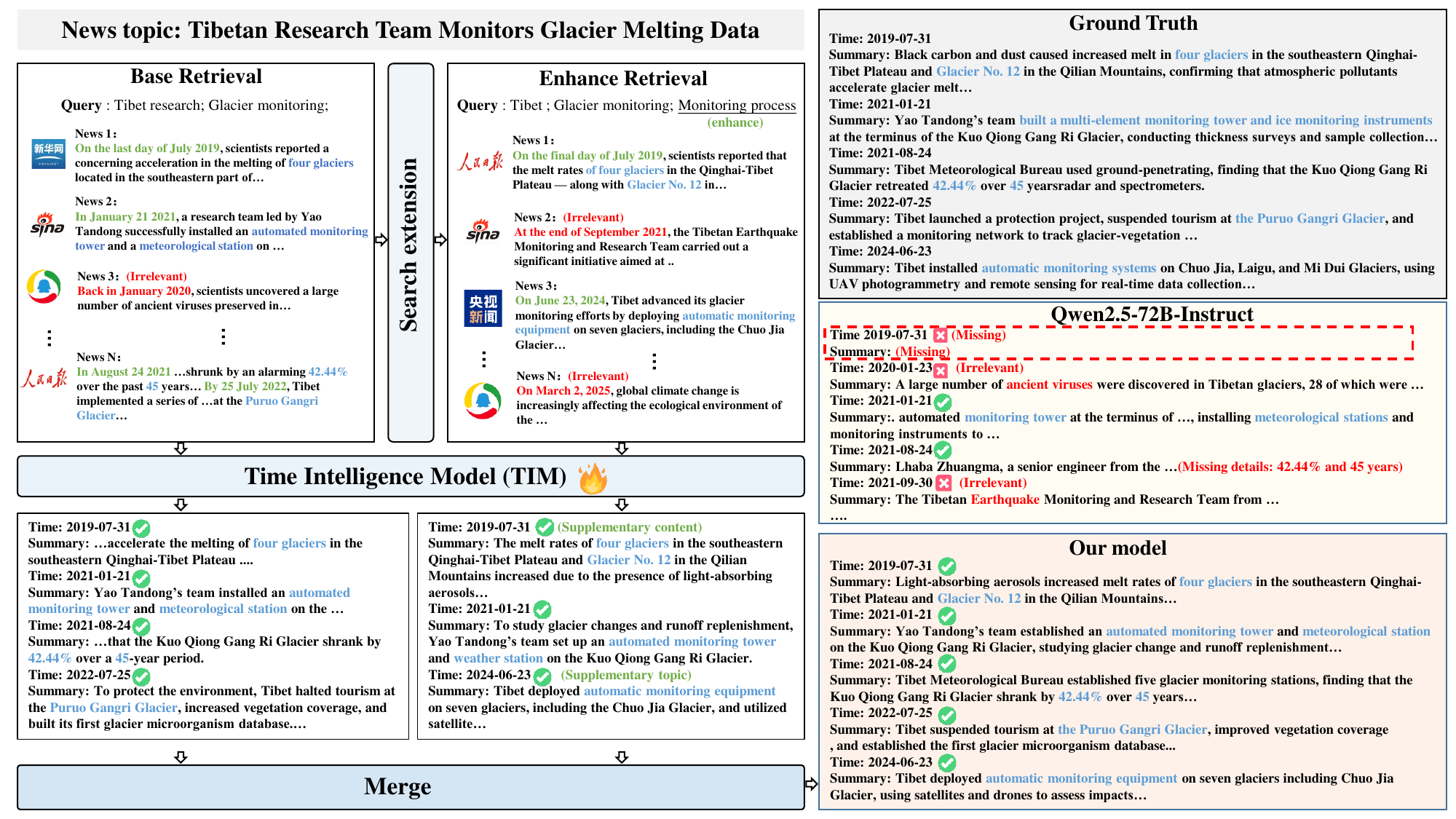}
  \caption{Overview of the TLS framework for the topic “Tibetan Research Team Monitors Glacier Melting Data.” The process starts with Base Retrieval using simple queries to gather initial documents. To enhance coverage, Search Extension reformulates the query (e.g., focusing on “monitoring process”) to retrieve supplementary evidence. Both sets of documents are processed by the TIM, which filters out irrelevant content, selects salient sentences, and temporally aligns them with associated timestamps. TIM also identifies supplementary content (in green) to enrich the TLS. Finally, the Merge step integrates all time-stamped summaries into a coherent TLS.}
  \label{fig:2}
\end{figure*}

\section{Related Work}
\subsection{Timeline summarization}
Before the advent of LLMs, TLS tasks were typically approached through three main paradigms: Direct Summarization \cite{martschat2018temporally,yu2021multi}, Date-wise Methods \cite{steen2019abstractive,ghalandari2020examining}, and Event Detection \cite{la2021summarize,yu2021multi}. With the rise of large language models (LLMs), several recent works have explored the use of LLMs for TLS. For instance, \cite{hu2024moments} proposed LLM-based event detection and incremental clustering to improve interpretability in a bounded context; \cite{qorib2025just} combined summarization and clustering with self-reflection to highlight salient events; and \cite{zhang2024dtels} introduced dynamic granularity timelines based on user instructions within specific domains. 
However, these methods are primarily designed for domain-specific or curated datasets, which limits their generalization capabilities and practical applicability. In this context, \cite{wu2025unfolding} first proposed the concept of open-domain TLS. They integrated Retrieval-Augmented Generation (RAG) with TLS by retrieving news through a search engine and subsequently employing a general LLM for summarization.
But they still face several limitations common to general-purpose LLMs when applied to open-domain inputs. These issues primarily arise from the absence of task-specific knowledge. Therefore, there is a growing need to develop a large timeline intelligence model that can estimate event salience, enforce temporal grounding, and maintain topic coherence—so as to generate accurate and informative timelines across diverse and dynamic real-world topics.
\subsection{Timeline summarization Dataset}
Training a large timeline intelligence model necessitates a sufficiently diverse and large TLS dataset. However, existing datasets in both English \cite{binh2013predicting, tran2015timeline} and Chinese \cite{mao2021cntls, wu2025unfolding} are notably limited in scale. As shown in Table \ref{tab:1}, most datasets only contain fewer than 100 topics and timelines, which is inadequate for effective model training. Additionally, these datasets often contain information from specific domains, hindering the validated of the model's generalization capabilities. Therefore, there is an urgent need to construct a large-scale, high-quality dataset for timeline summarization tasks.


\section{Preliminary}

\subsection{Task Definition}
The TLS task aims to generate a timeline that reflects the evolution of topics. It takes a user-specified news query as input and output a coherent, timestamped narrative of the topic's development. Let the user-specified news query be denoted as \( q \). Using a web search engine, such as Bing, we can retrieve a corresponding set of news articles, denoted as \( \mathcal{A} = \{A_1, A_2, \dots, A_n\} \), where \( n \) represents the number of news returned by the search engine. Based on the query \( q \) and the news article set \( A \), a LLM is employed to generate the timeline summary for the given news query \( q \), denoted as \( \mathcal{S}\):

\begin{equation}
\mathcal{S} = \text{LLM}_{\text{gen}}(q, A)
\end{equation}

Here, \( \mathcal{S}= \{S_1, S_2, \dots, S_l\} \) represents a set of timestamps, containing a total of \( l \) time points. Each time point is associated with a summary of length \( k \) that is relevant to the news query. All time points \( l \) are derived from the article set \( A \). The term \( \text{LLM}_{\text{gen}} \) refers to the LLM responsible for generating the TLS for the given news query.


\subsection{TLS framework}
The overall TLS framework is illustrated in Figure~\ref{fig:2}. Given a news topic query \( q \), an initial news article set \( \mathcal{A} \) is retrieved from a search engine. As this base retrieval often lacks sufficient temporal coverage and contextual richness, we introduce a \textit{search extension}. Starting from \( \mathcal{A} \), a LLM generates clarification questions to uncover implicit information needs. The extracted keywords are then reformulated into new queries to retrieve additional documents. A relevance model~\cite{xiao2024c} is used to rerank and filter these results, producing a refined set \( \mathcal{A}_{\text{base}} \). To further enhance temporal completeness and background diversity, a query generation is applied, yielding an enhanced set \( \mathcal{A}_{\text{enhanced}} \). 
Then, the TIM is applied to both \( A_{\text{base}} \) and \( A_{\text{enhanced}} \) to obtain \( \mathcal{S}_{\text{base}} \) and \( \mathcal{S}_{\text{enhance}} \). Finally, we employ a general LLM to merge the two TLS into the topic TLS \( \mathcal{S}_{\text{merge}} \). The TIM model is a trainable large-scale model specifically designed for the TLS task, while the merge module without requiring additional training (see Section 6.3.1).

\section{TLS Dataset}
We construct a large-scale TLS dataset, with each instance refined through automated processing and manual correction. Compared to existing datasets like T17 \cite{binh2013predicting} and Crisis \cite{tran2015timeline}, which are limited in temporal span and topic diversity, our dataset offers broader coverage, longer timeframes, and more diverse sources—making it better suited for building a large timeline intelligence model for TLS.
\subsection{Dataset Collection}
\begin{figure}[t]
  \centering
  \includegraphics[width=\linewidth]{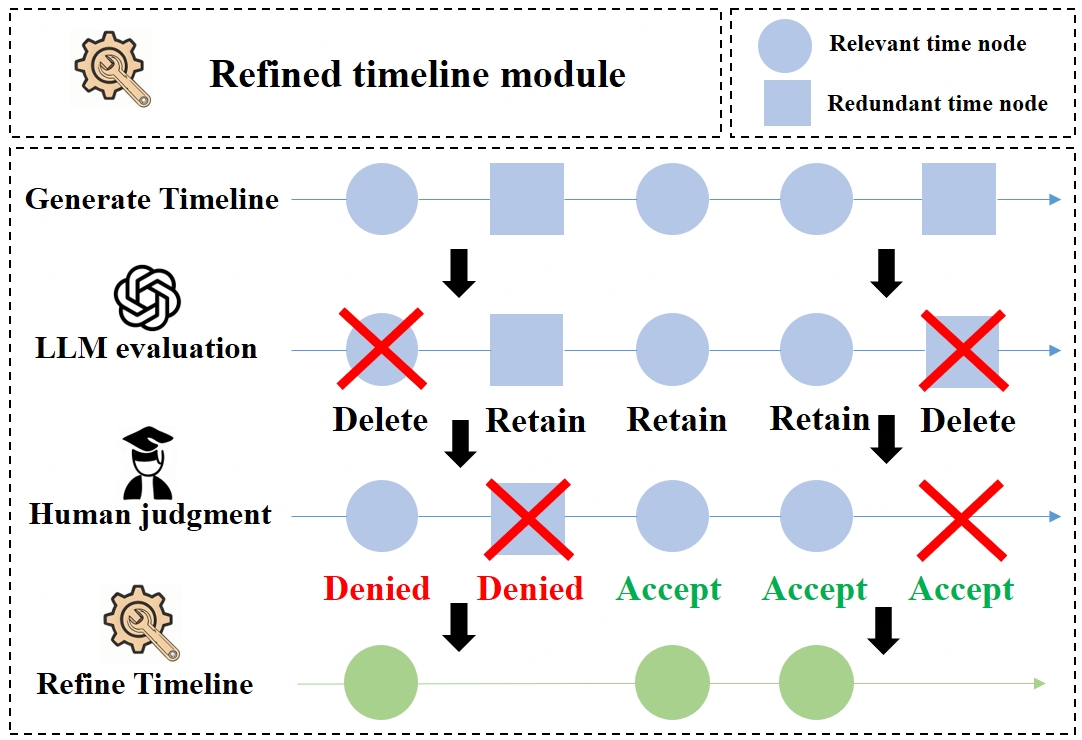}
  \caption{Illustration of TLS refinement. For each timestamp, the output of LLM is reviewed and voted on by human experts to ensure the accuracy and reliability.}
  \label{fig:2.1}
\end{figure}

Traditionally, the collection of news topics relied on manual search and summarization. By leveraging advances in LLMs and web search tools \cite{liu2024deepseek}, we automate this process through LLM-guided web searches across 12 domains (e.g., politics, sports) from January 2024 to January 2025, segmented by month. After manual filtering, we compile a diverse set of 1,189 topics, denoted as \( Q \). We then employ the framework described in Section 3.2, utilizing GPT-4o \footnote{\url{https://openai.com/index/hello-gpt-4o/}} for generating a basic timeline and for filtering irrelevant information through the model's self-reflection capabilities. Finally, no fewer than five artificial experts review and refine the output to ensure accuracy. Fleiss' Kappa \cite{fleiss1971measuring} is employed to measure inter-annotator consistency. For samples with a Kappa score below 0.8, further discussions are conducted until consensus is reached. The entire process is illustrated in Figure \ref{fig:2.1}. Ultimately, we obtain the first large-scale, high-quality TLS dataset (TLS-I).


\subsection{Dataset Statistics}
The TLS-I dataset contains 1,189 news topics with more than 100,000 collected source news. Detailed dataset statistics are presented in Table \ref{tab:1}. Notably, our dataset contains the largest number of topics and timeline summaries among existing TLS datasets, providing sufficient scale to support large timeline intelligence model training. Each topic contains three timelines: \( \mathcal{S}_{\text{base}} \),  \( \mathcal{S}_{\text{enhance}} \), and \( \mathcal{S}_{\text{merged}} \). This structure enables researchers to evaluate model performance across various research directions, including Retrieval-Augmented Generation (RAG), event extraction, and the identification and filtering of redundant information.

\begin{figure}[t]
  \includegraphics[width=\columnwidth]{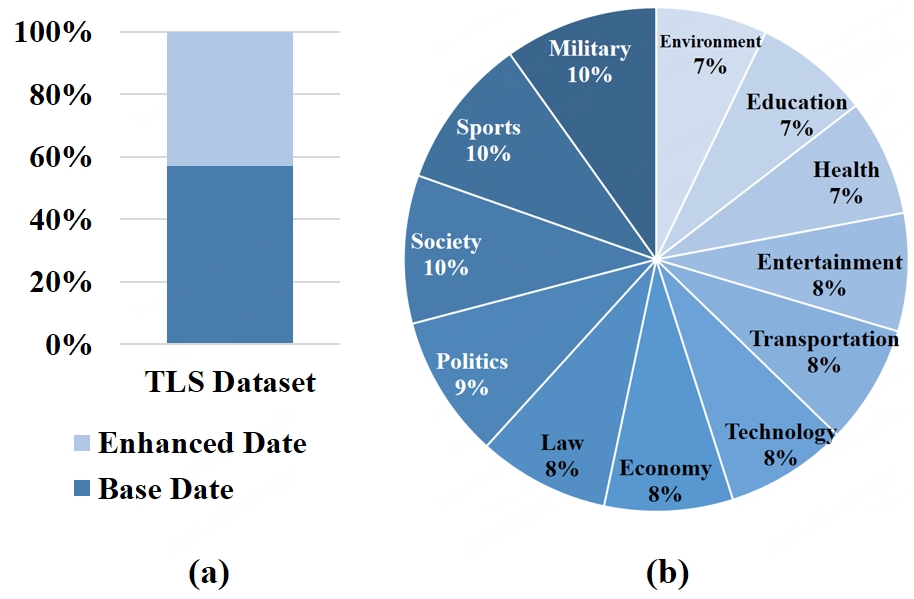}
  \caption{(a) Illustration of the proportion of basic and enhanced timestamps within the TLS of each news topic in the dataset. (b) The distribution of news topics across different domains in the entire dataset.}
  \label{fig:2.5-m}
\end{figure}

Figure \ref{fig:2.5-m}(a) depicts the proportion of basic versus enhanced timestamps within each merged TLS. The ratio of temporal nodes in basic to enhanced timelines is approximately 6:4. This indicates that relying solely on basic searches, without conducting extended searches, may lead to incomplete retrievals. In this case, even manual efforts are fail to capture the complete timeline of events. This underscores the necessity of incorporating search extension within our framework. Additionally, as depicted in Figure \ref{fig:2.5-m}(b), our dataset encompasses a wide range of domains with a balanced distribution. This means that models trained on this dataset are expected to have good generalization ability.

\section{Progressive Optimization Strategy}
To enable the model have the ability to understand of topic relevance and topic evolution, we propose a Progressive Optimization Strategy. This strategy consists of two complementary optimization stages: the first stage employ instruction tuning to enhance the model’s ability of topic identification and understanding; the second stage incorporates a dual-alignment reward learning that encourages the model to learn topic evolution patterns from both temporal and semantic perspectives.

\subsection{Instruction Tuning}
To enable the model to identify topic-related information, we employ instruction tuning to fine-tune general LLMs. The effectiveness of instruction tuning hinges on constructing a training set with an appropriate level of difficulty. Therefore, we propose a topic-aware sampling strategy that acquires both high and low topic relevance samples, aiming to ensure that the topic relevance of the final dataset approximates a normal distribution. For each news query, we sample 10 highly relevant documents and 10 low-relevance documents from an average of 1,053 candidate documents using a combination of automatic ranking and manual filtering. The high topic relevance samples consist exclusively of content strongly related to the news topic, allowing the model to rapidly develop summarization skills. Conversely, the low topic relevance samples enhances the model's ability to discriminate and perceive topic relevance, thereby improving the robustness and accuracy of information filtering.

Specifically, we define $\mathcal{D}_{\text{high}} = \{(a_{base}^{\text{i}}, s_{base}^{\text{i}})\}$ as the set of samples with high topic relevance. Similarly, we define 
$\mathcal{D}_{\text{low}} = \{(a_{enhance}^{\text{i}}, s_{enhance}^{\text{i}})\}$ as the set of samples with low topic relevance.

The objective function is defined as:
\begin{equation}
\begin{aligned}
\mathcal{L}_{\text{topic-aware}} = \sigma(\beta) \cdot \mathbb{E}_{(x, y) \sim \mathcal{D}_{\text{high}}} \left[ \ell(f_\theta(x), y) \right] \\
+ \left(1 - \sigma(\beta)\right) \cdot \mathbb{E}_{(x, y) \sim \mathcal{D}_{\text{low}}} \left[ \ell(f_\theta(x), y) \right]
\end{aligned}
\end{equation}

Here, \( \ell(\cdot) \) denotes the cross-entropy loss function. The function \( \sigma(\cdot) \) refers to the sigmoid function, and \( \beta \in \mathbb{R} \) is a learnable parameter. Specifically, we define \( \alpha = \sigma(\beta) \), which is used to control the relative contribution of high- and low-relevance samples to the total loss. This formulation allows the model to dynamically learn an optimal weighting strategy during training. This loss formulation encourages the model to distinguish between topic-relevant and irrelevant content, thereby enhancing its topic awareness. By jointly leveraging high- and low-relevance samples with a learnable weighting mechanism, the model is guided to focus on salient information while suppressing off-topic noise. This improves the overall coherence and topic consistency of the generated timeline summaries. We refer to the model trained using this approach as TIM (Standard).

\subsection{Dual-alignment Reward Learning}
Furthermore, to enhance the model’s ability of topic evolution patterns capturing, we design a dual-alignment reward learning approach. Inspired by the principles of Direct Preference Optimization (DPO) \cite{rafailov2023direct}, this method utilizes Alignment F1 \cite{martschat2018temporally} as a comprehensive evaluation metric and constructs positive and partially aligned negative samples. This encourages the model to optimize not only semantic alignment with reference summaries but also temporal alignment of event timestamps, thereby improving the logical coherence and temporal consistency of TLS. Preference alignment is achieved using the following:

\begin{align}
\mathcal{L}_{\text{dual}}(a_i, s_i^+, s_i^-) &= -\log \sigma\left( \beta \cdot \left( \log \pi\theta(s_i^+|a_i) \right. \right. \nonumber \\
& \quad \left. \left. - \log p_\theta(s_i^-|a_i) \right) \right)
\end{align}

where $y_i^+$ is the TLS with the highest score, $y_i^-$ is the TLS with the lowest score. $\beta$ controls the degree of constraint
and we refer to the model trained using this approach as TIM (Pro).

\begin{table*}[t]
\centering
\resizebox{\textwidth}{!}{%
\begin{tabular}{cccccccc}
\hline
\multirow{3}{*}{\textbf{Model}} & \multicolumn{4}{c}{\textbf{Alignment F1}}                                         & \multicolumn{2}{c}{\multirow{2}{*}{\textbf{Date F1}}} & \multirow{3}{*}{\textbf{\begin{tabular}[c]{@{}c@{}}the volume \\ of refusal\end{tabular}}} \\
                                & \multicolumn{2}{c}{\textbf{R-1}}      & \multicolumn{2}{c}{\textbf{R-2}}      & \multicolumn{2}{c}{}                                  &                                                                                            \\ \cline{2-7}
                                & \textbf{full}  & \textbf{non-refusal} & \textbf{full}  & \textbf{non-refusal} & \textbf{full}          & \textbf{non-refusal}         &                                                                                            \\ \cline{1-7}
\textbf{Qwen2.5-7B-Instruct}    & 0.213          & 0.200                & 0.092          & 0.089                & 0.529                  & 0.527                        & -                                                                                          \\
\textbf{Qwen2.5-14B-Instruct}   & 0.263          & 0.266                & 0.125          & 0.153                & 0.594                  & 0.610                        & -                                                                                          \\
\textbf{Qwen2.5-32B-Instruct}   & 0.275          & 0.279                & 0.133          & 0.146                & 0.604                  & 0.604                        & -                                                                                          \\
\textbf{Qwen2.5-72B-Instruct}   & 0.249          & 0.258                & 0.142          & 0.155                & 0.522                  & 0.529                        & -                                                                                          \\
\textbf{Llama3.3-70B-Instruct}  & 0.217          & 0.220                & 0.090          & 0.094                & 0.610                  & 0.612                        & -                                                                                          \\
\textbf{GPT-4o}                 & -              & 0.279                & -              & 0.168                & -                      & 0.645                        & 33                                                                                         \\
\textbf{Qwen2.5-Max-2025-01-25} & -              & 0.282                & -              & 0.157                & -                      & 0.614                        & 18                                                                                         \\
\textbf{Doubao-1.5-Pro}         & -              & 0.322                & -              & 0.171                & -                      & 0.597                        & 19                                                                                         \\
\textbf{DeepSeek-V3-0326}       & -              & 0.336                & -              & 0.209                & -                      & 0.649                        & 20                                                                                         \\\hline
\textbf{TIM-7B (Standard)}           & 0.338          & 0.346                & 0.168          & 0.200                & 0.623                  & 0.641                        & -                                                                                          \\
\textbf{TIM-14B (Standard)}          & 0.382          & 0.389                & 0.222          & 0.227                & 0.652                  & 0.660                        & -                                                                                          \\\hline
\textbf{TIM-7B (Pro)}       & 0.368          & 0.371                & 0.178          & 0.201                & 0.642                  & 0.636                        & -                                                                                          \\
\textbf{TIM-14B (Pro)}      & \textbf{0.410} & \textbf{0.413}       & \textbf{0.231} & \textbf{0.251}       & \textbf{0.653}         & \textbf{0.674}               & -                                                                                          \\ \hline
\end{tabular}%
}
\caption{The results of various models on our test set are presented, with metrics \textbf{bolding} the best performance. For topics with politically sensitive content, API models refuse to answer, so we separate the results into the full dataset and the non-refusal dataset. The last column shows the number of refusals by each API model.}\label{tab:2}
\end{table*}

\section{Experiments}

\subsection{Experimental Settings}

\textbf{Compared Models:} In our experiments, we compare the latest state-of-the-art models, both open-source and proprietary, including the Qwen2.5 (7B, 14B, 32B, 72B) instruct models \cite{yang2024qwen2}, Qwen2.5-Max-2025-01-25\footnote{\url{https://qwenlm.github.io/blog/qwen2.5-max/}}, Llama3.3-70B-Instruct \cite{grattafiori2024llama}, GPT-4o, Doubao-1.5-Pro, and DeepSeek-V3-0326\footnote{\url{https://api-docs.deepseek.com/zh-cn/news/news250325}}.In addition, we also evaluated large reasoning models \cite{el2025competitive}. However, due to API policy restrictions, most of the requests failed. Refer to the Appendix \ref{app:3} for detailed information.

\noindent\textbf{Implementation Details:} For TLS data collection, we leverage GPT-4o for all LLM-involved components, including self-questioning, keyword extraction, TLS generation, merging, and refinement. For document retrieval, we use Bing Search\footnote{\url{https://www.microsoft.com/en-us/bing/apis/bing-web-search-api}}, followed by reranking using the \texttt{bge-reranker-v2-minicpm-layerwise} model \cite{chen2024bge} with the 28th layer and top-K = 10 outputs to ensure the selection of high-relevance, high-quality content. Detailed prompt templates used throughout the data generation pipeline are provided in Appendix \ref{app:1}.

For model training, we conduct full-parameter supervised fine-tuning on Qwen2.5-7B-Instruct and Qwen2.5-14B-Instruct using 8 A800 GPUs via the LLAMA-Factory framework \cite{zheng2024llamafactory}, with a learning rate of 5e-6, batch size of 64, and 3 epochs. Based on the resulting TIM (Standard) models, we further optimize TIM (Pro) models using OpenRLHF \cite{hu2024openrlhf}, adopting a learning rate of 5e-7, batch size of 128, and training for 2 epochs.
During inference, we replace GPT-4o with our TIM model for TLS generation, and use Qwen2.5-32B-Instruct as the merge model. LLMs are deployed locally using the vLLM framework \cite{kwon2023efficient}. All other components and configurations remain unchanged.

\noindent\textbf{Evaluation Protocols:} To ensure fairness, we utilized 132 news queries not present in the dataset to validate the capabilities of our timeline intelligence model and other LLMs. These queries span 12 domains to comprehensively evaluate the model's generalizability. Consistent with the data collection process, the final results were assessed by at least five human experts to obtain the evaluation metrics.

\begin{figure*}[t]
  \centering
  \includegraphics[width=\linewidth]{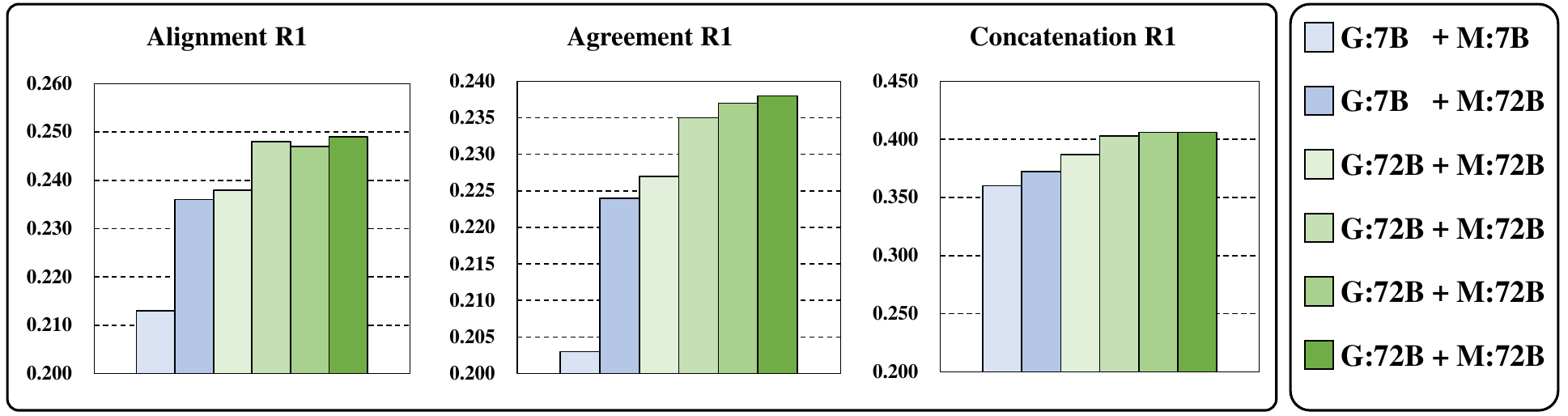}
  \caption{The figure shows the ROUGE-1 results for various metrics. "\textbf{G}" stands for Generation, and "\textbf{M}" stands for Merge. For example, "G:7B + M:72B" indicates that the TLS generation model is Qwen2.5-7B-Instruct, and the TLS merging model is Qwen2.5-72B-Instruct.}
  \label{fig:3}
  \vspace{-10pt}
\end{figure*}



\noindent\textbf{Evaluation Metrics:} We adopt the evaluation metrics proposed by \cite{martschat2017improving,martschat2018temporally}, which combine ROUGE and DATE to assess both writing quality and temporal alignment in TLS. \textit{Concatenation F1}: Computes the ROUGE score between the generated and groundtruth TLS. \textit{Agreement F1}: Computes the ROUGE score between the generated timeline and the groundtruth TLS for matching dates. \textit{Alignment F1}: Calculates the ROUGE score while also considering the alignment of dates between the generated and groundtruth TLS. \textit{Date F1}: Compares the dates of the generated TLS with the groundtruth TLS.

\begin{table*}[t]
\setlength{\abovecaptionskip}{0.05cm}
\setlength{\belowcaptionskip}{0cm}
\centering
\resizebox{\textwidth}{!}{%
\begin{tabular}{ccccccc}
\hline
\multirow{2}{*}{\textbf{Model}} & \multicolumn{2}{c}{\textbf{Base Alignment F1}} & \multirow{2}{*}{\textbf{Date F1}} & \multicolumn{2}{c}{\textbf{Enhance Alignment F1}} & \multirow{2}{*}{\textbf{Date F1}} \\ \cline{2-3} \cline{5-6}
                                & \textbf{R-1}               & \textbf{R-2}      &                                   & \textbf{R-1}                 & \textbf{R-2}       &                                   \\ \hline
\textbf{Qwen2.5-7B-Instruct}  & 0.261                      & 0.136             & 0.564                             & 0.276                        & 0.112              & 0.594                             \\
\textbf{TIM-7B (Pro)}         & $\mathbf{0.366}_{(\mathrm{+40.2\%})}$ & $\mathbf{0.188}_{(\mathrm{+38.2\%})}$ & $\mathbf{0.640}_{(\mathrm{+13.5\%})}$ & $\mathbf{0.332}_{(\mathrm{+20.3\%})}$ & $\mathbf{0.144}_{(\mathrm{+28.6\%})}$ & $\mathbf{0.641}_{(\mathrm{+7.9\%})}$ \\ \hline
\textbf{Qwen2.5-14B-Instruct} & 0.311                      & 0.174             & 0.641                             & 0.239                        & 0.111              & 0.617                             \\
\textbf{TIM-14B (Pro)}        & $\mathbf{0.425}_{(\mathrm{+36.7\%})}$ & $\mathbf{0.279}_{(\mathrm{+60.3\%})}$ & $\mathbf{0.696}_{(\mathrm{+8.6\%})}$ & $\mathbf{0.378}_{(\mathrm{+58.6\%})}$ & $\mathbf{0.189}_{(\mathrm{+70.3\%})}$ & $\mathbf{0.653}_{(\mathrm{+5.8\%})}$ \\ \hline
\end{tabular}%
}
\caption{Training preferences for models with different parameter sizes. The low-parameter (7B) model tends to focus on improving its summarization ability, while the high-parameter (14B) model prefers to enhance its information filtering capability.}\label{tab:3}
\vspace{-10pt}
\end{table*}


\subsection{Overall Performance}
The results are summarized in Table~\ref{tab:2}. Based on the results, we derive the following key findings:

\textbf{(1) General LLMs exhibit suboptimal performance on the TLS task, highlighting the necessity of task-specific knowledge.} Across all metrics, general LLMs—including GPT-4o, Qwen, and DeepSeek—struggle to effectively capture relevant, non-redundant information. This performance gap underscores the need of a large timeline intelligence tailored to the specific demands of the TLS task. Meanwhile, we find that all metrics improve as the model parameter size increase. This trend suggests that, given sufficient resources for training a larger model, improved results are anticipated.

\textbf{(2) TIM (Standard) demonstrates better performance than general LLMs.} This finding underscores the effectiveness of our instruction tuning in enhancing topic understanding and timeline quality. Compared with DeepSeek-V3-0326—a 685B general large language model—our 7B/14B TIM (Standard) model even achieves higher Alignment F1 scores. This improvement indicates that our model is better equipped to focus on salient events, maintain topic relevance, and minimize off-topic or redundant content.
These results demonstrate that task-specific knowledge can equip relatively lightweight models with strong alignment capabilities, even surpassing much larger foundation models in complex structured generation tasks like open-domain TLS.

\textbf{(3) TIM (Pro) achieves the state-of-the-art performance in the TLS task.} This underscores that understanding the basic laws of topic
evolution is important for TLS task. Through the dual-alignment reward learning, TIM gains deeper insights into topic evolution, enabling it to more effectively eliminate irrelevant time nodes. Consequently, TIM (Pro)-7B/14B demonstrates consistent improvements across all evaluation metrics. Notably, TIM-14B (Pro) achieves an average performance improvement of over 14\% compared to DeepSeek-V3-0326. This suggests that TIM outperforms general LLMs in TLS tasks, particularly in open domains, demonstrating its potential for practical applications.

\subsection{Analysis and Discussion}
\subsubsection{Generation vs. Merging}
Figure \ref{fig:3} shows that larger-parameter generation models (green bars) achieve lower performance metrics than their smaller-parameter counterparts (blue bars). Furthermore, when the merging model's parameter size exceeds 32B, there is no improvement in its performance. This observation motivates our experimental design: we concentrate on enhancing the smaller generation models, while retaining the un-finetuned 32B model for the merging stage. This is because that the generation model must handle large‑scale text: long‑form summarization, retrieval‑augmented generation, and extensive information filtering. By contrast, the merging model only needs to integrate two structured TLS outputs. Therefore, given the need for a specific solution and the limitations of existing general models, training a dedicated generative model offers a more effective approach than model merging.

\begin{table}[t]
\setlength{\abovecaptionskip}{0.05cm}
\setlength{\belowcaptionskip}{0cm}
\centering
\resizebox{\linewidth}{!}{%
\begin{tabular}{lcccc}
\hline
\textbf{Model}                     & \multicolumn{1}{l}{\textbf{Concat R-2}} & \multicolumn{1}{l}{\textbf{Agree R-2}} & \multicolumn{1}{r}{\textbf{Align R-2}} & \multicolumn{1}{l}{\textbf{Date F1}} \\ \hline
\textbf{ w/ RS}            & 0.217                                   & 0.145                                  & 0.152                                  & 0.593                                \\
\textbf{ w/ US}         & 0.243                                   & 0.156                                  & 0.163                                  & 0.590                                \\
\textbf{ w/ TS} & \textbf{0.265}                          & \textbf{0.161}                         & \textbf{0.168}                         & \textbf{0.641}                       \\ \hline
\end{tabular}
}
\caption{The table presents experimental results of different sampling strategies based on the Qwen2.5-7B-Instruct model. Our topic-aware sampling (\textbf{TS}) consistently outperforms rejection sampling (\textbf{RS}) and uniform sampling (\textbf{US}) across all evaluation metrics.}
\label{tab:4}
\vspace{-15pt}
\end{table}

\subsubsection{Training-Free vs. Training}
As shown in Table \ref{tab:3}, we evaluate both Qwen·2.5-7B-Instruct (7B) and Qwen·2.5-14B-Instruct (14B) under our proposed progressive optimization strategy. The 7B model demonstrates the most notable improvement under the base TLS setting, benefiting from our topic-aware sample selection strategy, which helps smaller models focus on extracting topic-relevant and temporally coherent content. In contrast, the 14B model—already strong in instruction-following and summarization tasks (Table \ref{tab:2})—achieves greater gains under the enhanced TLS setting. This is largely due to its superior capacity to handle information redundancy and leverage richer contextual signals in the enhanced documents, which aligns well with the goals of our dual-alignment reward learning. Overall, the results validate the effectiveness of our proposed strategies and further emphasize that filtering off-topic and temporally inconsistent content remains a key challenge, particularly in noisy scenarios.

\subsubsection{Ablation Study}
Table \ref{tab:4} presents the results of our ablation study. Rejection sampling refers to training the model using only news that are highly relevant to the target topic, which limits the model’s ability to capture the broader context and the progression of events. Uniform sampling, on the other hand, involves training the model on news regardless of their topic relevance, which leads the model to focus solely on information extraction while neglecting topic evolution. This degrades the Date F1 score. In contrast, our method enables the model to not only maintain topic awareness but also effectively capture the temporal dynamics of topic development. Through this approach, the model gradually learns to recognize the patterns of topic evolution, allowing it to discard information at time points that do not align with the topic's evolution. Consequently, this leads to improved performance across all evaluation metrics.

\subsubsection{Case Study}
The right side of Figure \ref{fig:2} presents the summarization results of "Tibetan Research Team Monitors Glacier Melting Data" in an open-domain context. We observe that general LLM models often lose temporal timestamps and generate redundant, irrelevant timestamps. Compared to the ground truth, the Qwen2.5-72B-Instruct model fails to capture the key event on 2019-07-31 and generates several redundant events that do not appear on the reference timeline. Additionally, some summaries are unrelated to the news topic, with specific instances detailed in Appendix \ref{app:2}. In contrast, our model successfully identifies all relevant events. This improvement can be attributed to our model's capacity to understand topic relevance and the evolutionary patterns of topics. Consequently, our model demonstrates superior summarization performance.
\subsection{Conclusions}
This paper introduces a progressive optimization strategy for TLS, which enhances both topic relevance and temporal coherence in model outputs. To support this, we construct and release the first large-scale TLS dataset, TLS-\uppercase\expandafter{\romannumeral1}, covering over 1,000 topics and 100,000 news articles. Leveraging topic-aware sampling and dual-alignment reward learning, we develop the first TLS-oriented LLM, TIM, which achieves state-of-the-art performance on open-domain TLS tasks. Future work will explore LLM acceleration techniques to improve inference speed over large-scale inputs while maintaining high summarization quality.

\section*{Limitations}
Although this study has made progress in the field of TLS, by proposing an innovative training method and releasing the first large-scale, trainable Chinese theme TLS dataset, there are still some limitations: (1) Due to resource constraints, we only implemented 7B and 14B models; however, the experiments have already demonstrated strong scalability. We believe that even better performance can be achieved with larger-scale models. (2) This study does not involve extensive design of the search engine itself. Incorporating a more advanced search engine framework could yield cleaner and more relevant retrieved documents.
\section*{Ethical Concerns}
All data used in this study were collected from publicly available sources and do not include any non-public information. Any personally identifiable information has been anonymized to prevent the identification of individuals. The dataset is used solely for academic research and not for commercial purposes.

\bibliography{custom}
\nocite{*}
\appendix
\section{Appendix}

\subsection{LLM prompt}
\label{app:1}
\begin{figure*}[t]
  \centering
  \includegraphics[width=\linewidth]{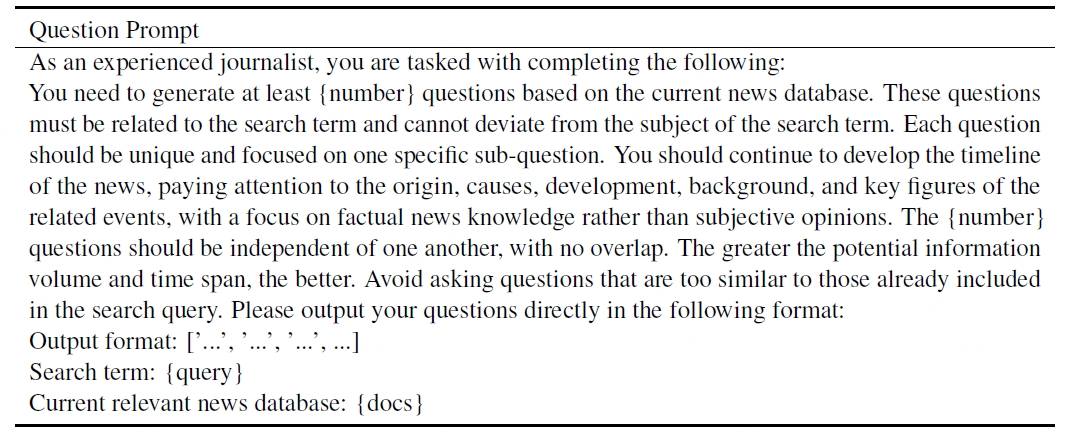}
  \caption{The prompt for self-questioning}
  \label{fig:question}

\end{figure*}

\begin{figure*}[]
  \centering
  \includegraphics[width=\linewidth]{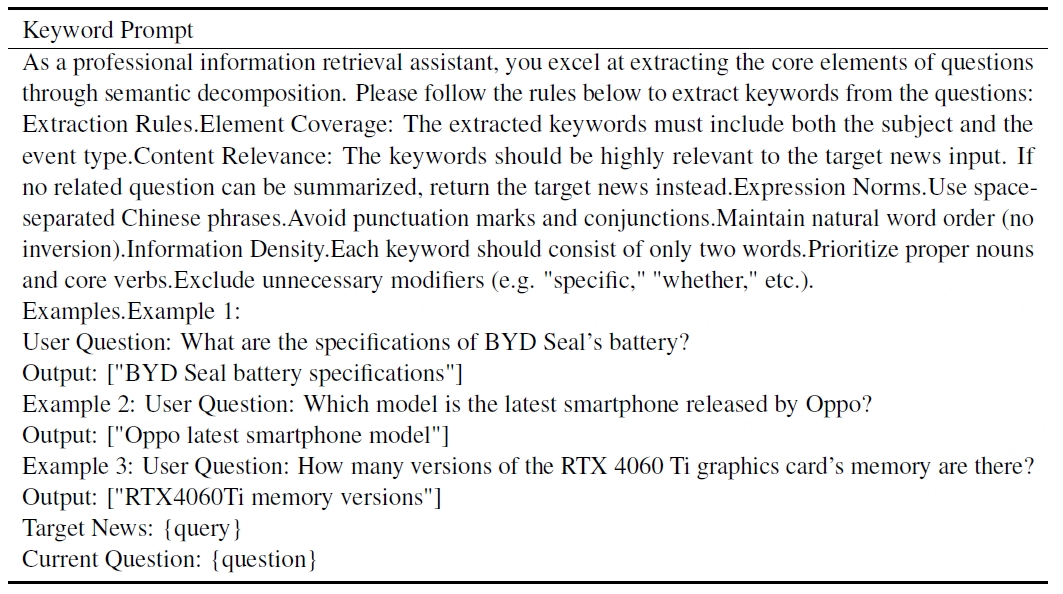}
  \caption{The prompt for keyword generation}
  \label{fig:key}
\end{figure*}

\begin{figure*}[]
  \centering
  \includegraphics[width=\linewidth]{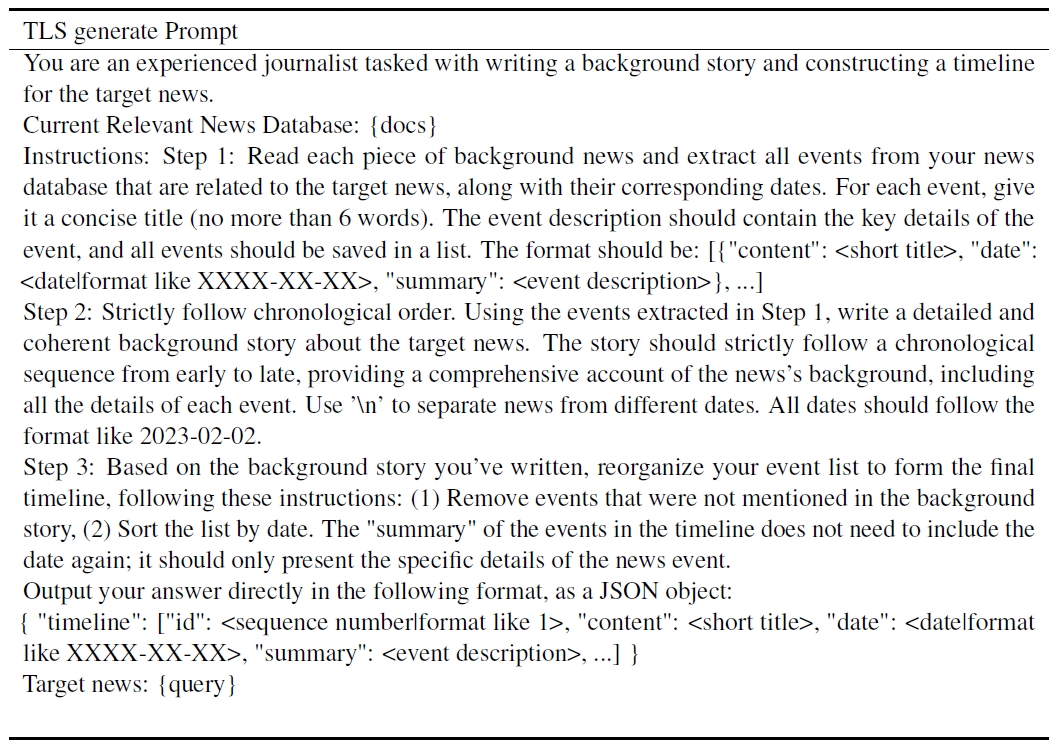}
  \caption{The prompt for TLS generation}
  \label{fig:gen}
\end{figure*}

\begin{figure*}[]
  \centering
  \includegraphics[width=\linewidth]{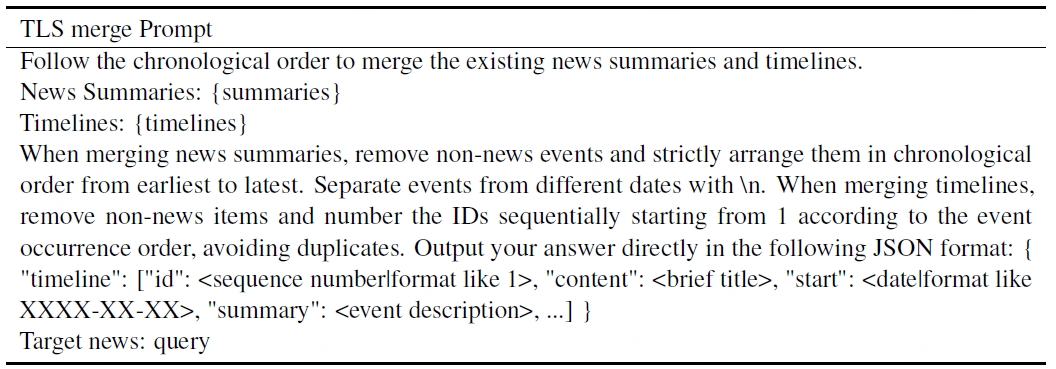}
  \caption{The prompt for TLS merging}
  \label{fig:merge}
\end{figure*}

\begin{figure*}[]
  \centering
  \includegraphics[width=\linewidth]{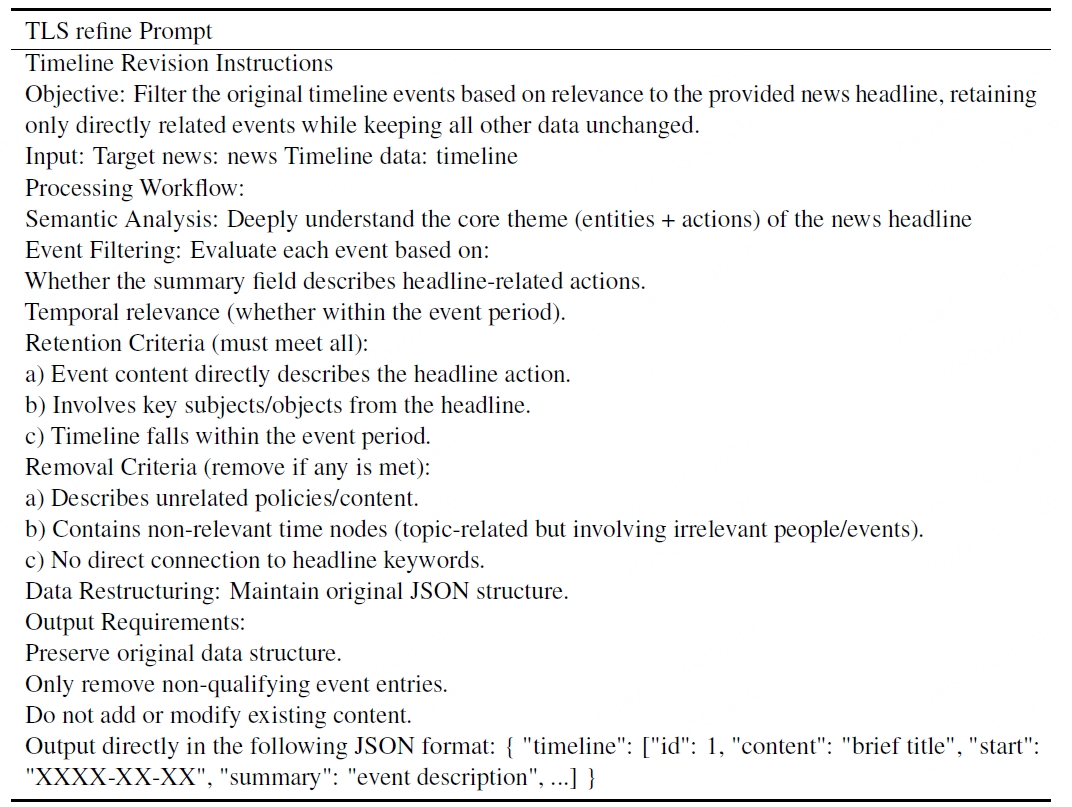}
  \caption{The prompt for TLS refinement}
  \label{fig:refine}
\end{figure*}

This section presents all reference prompts related to inputs for LLMs. Prompts \ref{fig:question} and \ref{fig:key} correspond to self-asking and keyword generation, respectively, which are applied in the Search Extension Module. Prompt \ref{fig:gen} is used for TLS generation, while Prompt \ref{fig:merge} is employed for TLS merging. Prompt \ref{fig:refine} provides guidance for the refinement of each time node within the TLS.

\subsection{Case study}
\label{app:2}
\begin{figure*}[]
  \centering
  \includegraphics[width=\linewidth]{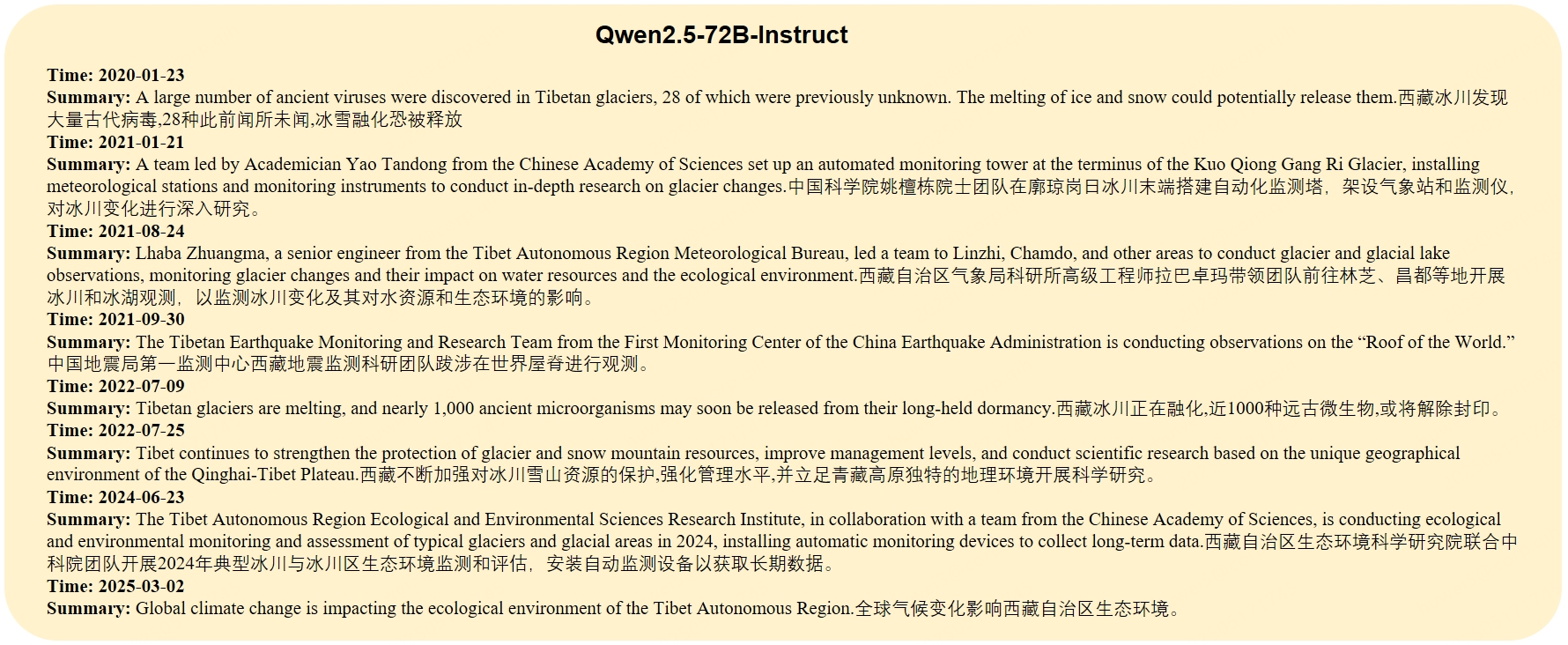}
  \caption{The TLS generated by Model Qwen2.5-72B-Instruct is illustrated in the figure above. The input news headline is \texttt{``Tibetan Research Team Monitors Glacier Melting Data.''}}
  \label{fig:case1}
\end{figure*}

\begin{figure*}[]
  \centering
  \includegraphics[width=\linewidth]{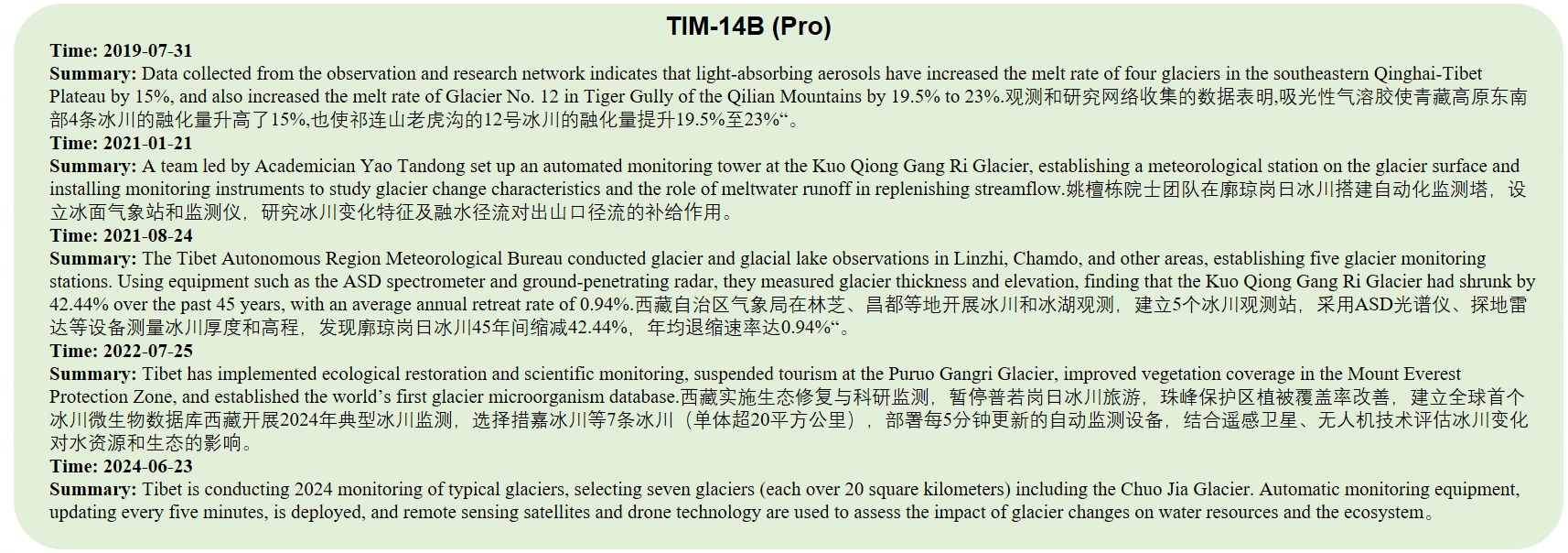}
  \caption{The TLS generated by TIM-14B (Pro) is illustrated in the figure above. The input news headline is \texttt{``Tibetan Research Team Monitors Glacier Melting Data.''}}
  \label{fig:case2}
\end{figure*}

\begin{figure*}[]
  \centering
  \includegraphics[width=\linewidth]{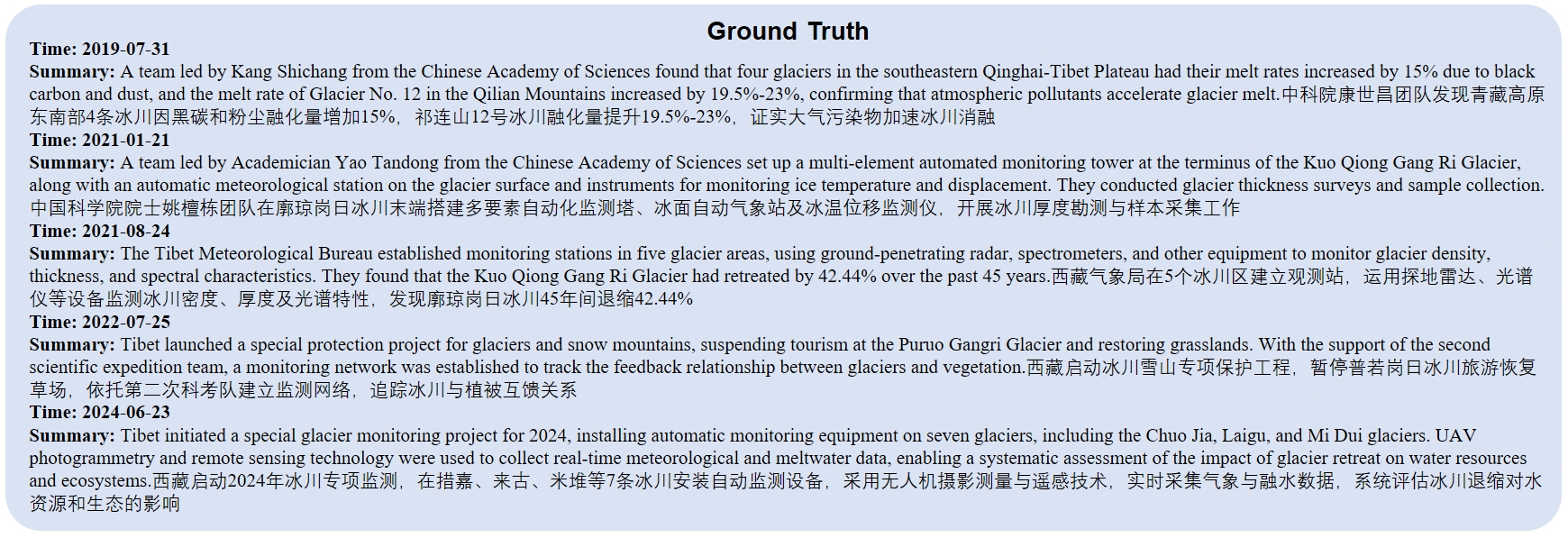}
  \caption{The groundtruth is shown in the figure above. The input news headline is \texttt{``Tibetan Research Team Monitors Glacier Melting Data.''}}
  \label{fig:case3}
\end{figure*}

Figures \ref{fig:case1}, \ref{fig:case2}, and \ref{fig:case3} represent the TLS generation results produced by different models. Each summary includes both Chinese and English versions for comparison. By referring to Figure \ref{fig:2}, one can identify specific time points where the summarized content is irrelevant to the actual news topic.

\subsection{Large reasoning model API policy}
\label{app:3}
We tested a range of large reasoning models, including Gemini 2.5 Pro\footnote{\url{https://ai.google.dev/gemini-api/docs/models\#gemini-2.5-pro-preview-05-06}}, DeepSeek-R1 \cite{guo2025deepseek}, Claude-3.7-sonnet-20250219\footnote{\url{https://docs.anthropic.com/en/api/overview}}, and GPT-o1\footnote{\url{https://platform.openai.com/docs/models/o1}}. However, due to the nature of the news queries and retrieved content—which often involve politically sensitive or ethically complex topics—many of these models refused to generate outputs as part of their internal reasoning processes, or triggered API-level safety restrictions. As a result, we were unable to obtain valid TLS outputs from these models, and thus their results are not reported in the Table \ref{tab:3}.

\end{CJK}
\end{document}